\documentclass{bmvc2k}

\usepackage{booktabs}
\usepackage{url}
\usepackage{amsmath}
\usepackage{amssymb}
\usepackage{graphicx}
\usepackage{amsfonts}

\title{Adversarial Color Enhancement:\\Generating Unrestricted Adversarial Images by Optimizing a Color Filter}

\addauthor{Zhengyu Zhao}{z.zhao@cs.ru.nl}{1}
\addauthor{Zhuoran Liu}{z.liu@cs.ru.nl}{1}
\addauthor{Martha Larson}{m.larson@cs.ru.nl}{1}

\addinstitution{
 Data Science Group,\\
 Institute for Computing and Information Sciences,\\
 Radboud University,\\
 Nijmegen, Netherlands
}
\runninghead{ZHAO, LIU, LARSON}{Adversarial Color Enhancement}

\begin{document}

\maketitle

\begin{abstract}
We introduce an approach that enhances images using a color filter in order to create adversarial effects, which fool neural networks into misclassification.
Our approach, Adversarial Color Enhancement (ACE), generates unrestricted adversarial images by optimizing the color filter via gradient descent.
The novelty of ACE is its incorporation of established practice for image enhancement in a transparent manner.
Experimental results validate the white-box adversarial strength and black-box transferability of ACE.
A range of examples demonstrates the perceptual quality of images that ACE produces.
ACE makes an important contribution to recent work that moves beyond $L_p$ imperceptibility and focuses on unrestricted adversarial modifications that yield large perceptible perturbations, but remain non-suspicious, to the human eye.
The future potential of filter-based adversaries is also explored in two directions: guiding ACE with common enhancement practices (e.g., Instagram filters) towards specific attractive image styles and adapting ACE to image semantics.
Code is available at~\url{https://github.com/ZhengyuZhao/ACE}.
% The future potential of color filter-based adversaries is demonstrated by two directions also explored in this paper: adapting ACE to image semantics and guiding ACE with common filters widely used online to enhance images.
\end{abstract}

%-------------------------------------------------------------------------
% In order to make the modification non-suspicious to a human observer, most related work emphasizes the imperceptibility of the additive perturbations, i.e., pair-wise similar between the adversarial image and its clean version, mainly with respect to $L_p$ distance. 
% However, despite being mostly modified from a clean starting point, recent work has demonstrated that an adversarial image does not really require a tight relationship with its clean version for ensuring modification to be non-suspicious.
% Therefore, unrestricted yet natural-looking adversarial images are instead designed.
\section{Introduction}
\label{sec:intro}
Despite the exceptional success of the Deep Neural Networks (DNNs), recent research has shown that they are remarkably susceptible to \textit{adversarial examples}~\cite{szegedy2013intriguing}, which are crafted to induce incorrect model predictions.
Adversarial image examples have been extensively studied in image classification~\cite{carlini2017towards,goodfellow2014explaining,joshi2019semantic,madry2017towards,moosavi2016deepfool,papernot2016limitations,xiao2018spatially}, and also explored in object detection~\cite{chen2018shapeshifter,zhao2019seeing}, semantic segmentation~\cite{arnab2018robustness,xie2017adversarial} and image retrieval~\cite{liu2019s,tolias2019targeted}.

A key property of adversarial images that makes them dangerous is that they cause decision conflicts between the model and human annotated labels in a way that is hardly recognizable to human~\cite{gilmer2018motivating,sharif2018suitability}.
Most conventional work on adversarial examples has focused on imperceptible additive perturbations, whereby imperceptibility is conventionally measured with the $L_p$ distance between the adversarial images and their clean versions~\cite{carlini2017towards,moosavi2016deepfool,papernot2016limitations}.
Later studies proposed to leverage more perception-aligned measurements~\cite{Croce_2019_ICCV,luo2018towards,wong2019wasserstein,zhang2020smooth,zhao2020towards} to address the well known insufficiency of naive $L_p$ norms as perceptual similarity metric~\cite{wang2004image}, 
but have still focused exclusively on imperceptible perturbations.

\begin{figure}[t]
\begin{center}
% \fbox{\rule{0pt}{3.2in} \rule{\textwidth}{0pt}}
  \includegraphics[width=\columnwidth]{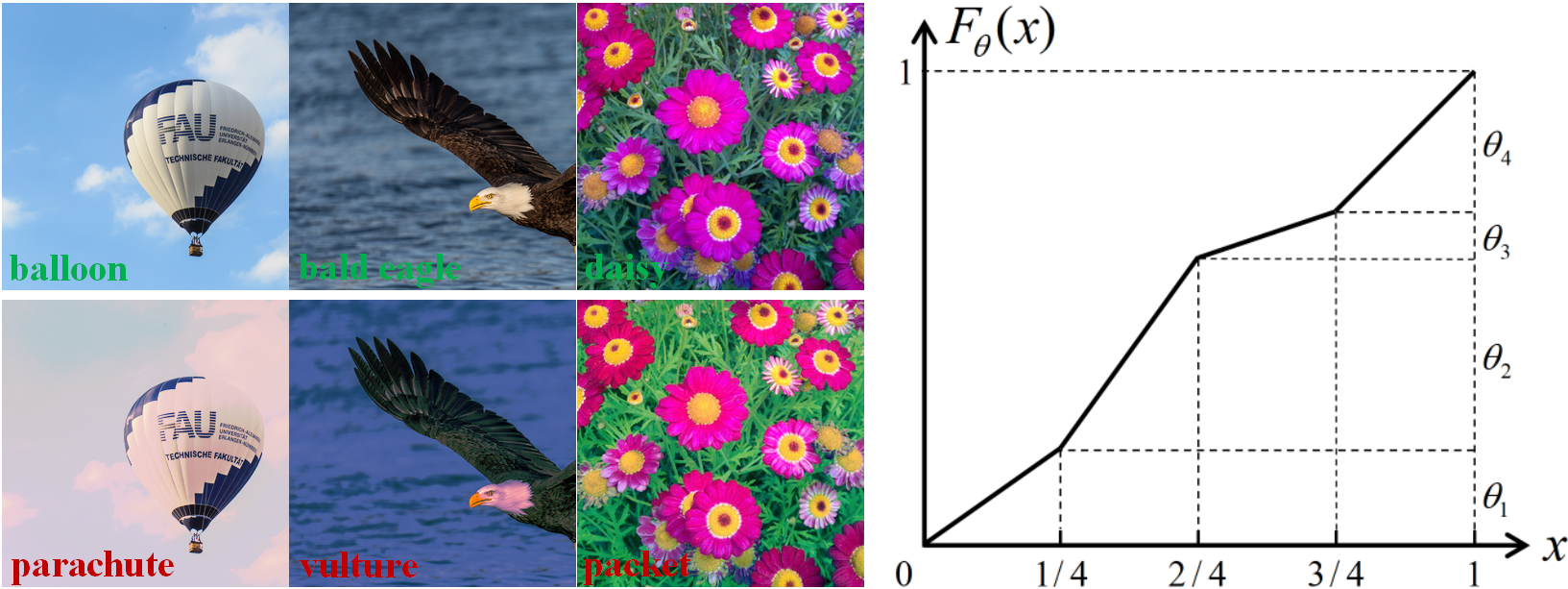}
\end{center}
% \vspace{-0.2cm}
   \caption{\textbf{Left:} Original Images (\textbf{top}) and their adversarial versions (\textbf{bottom}) generated by our Adversarial Color Enhancement (ACE). Additional examples can be found in our GitHub repository. \textbf{Right:} Illustration of the color filter adopted in ACE (here $K=4$ in Equation~\ref{filter}).}
\label{fig:examples}
% \vspace{-0.2cm}
\end{figure}

Recently, it has been pointed out that when small, imperceptible perturbations were originally introduced by~\cite{goodfellow2014explaining}, they were intended only to be an abstract, toy example for easy evaluation, and that actually it is hard to find a compelling example that requires imperceptibility in realistic security scenarios~\cite{gilmer2018motivating}.
In other words, imposing similarity with respect to an original, clean image is not necessary in real-world threat models.
For this reason, recent work has moved beyond small imperceptible perturbations, and started exploiting ``unrestricted adversarial examples''~\cite{brown2018unrestricted} that have natural looks even with large, visible perturbations, but remain non-suspicious to the human eye~\cite{bhattad2020Unrestricted,joshi2019semantic,engstrom2019exploring}.
In general, exploring new types of threat models beyond conventional imperceptible perturbations will provide a more comprehensive understanding of adversarial robustness of the DNNs~\cite{xiao2018spatially}.
And more importantly, relaxing the tight bound on perturbations has been shown to yield practically interesting properties, such as cross-model transferability for black-box adversaries applied in real-world scenarios~\cite{bhattad2020Unrestricted,shamsabadi2020colorfool}.

Building on these recent developments, we propose a new approach to generating unrestricted adversarial images using a color filter.
The approach, called Adversarial Color Enhancement (ACE), introduces non-suspicious perturbations, with minimal impact on image quality, as shown in Figure~\ref{fig:examples} (left).
Although previous work~\cite{choi2017geo} has pointed out that common enhancement practices (e.g., Instagram filters) can degrade the performance of the automatic geo-location estimation, until now, no research has focused on the optimization aspect of exploiting image filters to create adversarial images.
Our approach makes use of recent advances in automatic image retouching based on differentiable approximation of commonly-used image filters ~\cite{hu2018exposure,deng2018aesthetic}.
In sum, this paper makes the following contributions:
\begin{itemize}
  \setlength{\parskip}{0pt}
  \setlength{\itemsep}{0pt}
\item We explore the vulnerability of the DNNs to commonly-used image filters, and specifically propose Adversarial Color Enhancement (ACE), an approach to generating unrestricted adversarial images by optimizing a differentiable color filter. 
\item Experimental results demonstrate ACE achieves a better trade-off between the adversarial strength and perceptual quality of the filtered images than other state-of-the-art methods, implying a stronger black-box adversary for real-world applications.
\item We explore two potential ways to further improve ACE on image quality: 1) using widely-used enhancement practices (e.g., Instagram filters) as guidance to specified attractive image styles, and 2) leveraging regional semantic information.
\end{itemize}

\section{Related Work}
\noindent\textbf{Differentiable Image Filters.}
The state of the art for automatic photo retouching mainly uses supervised learning to determine editing parameters via gradient descent, in order to achieve specific image appearances.
Most approaches~\cite{chen2017fast,gharbi2017deep,isola2017image,yan2016automatic,zhu2017unpaired} utilize DNNs for the parameterization of the editing process, but inevitably 
they suffer from high computational cost, fixed image resolution, and more importantly, a lack of interpretability.
For this reason, some recent work~\cite{hu2018exposure,deng2018aesthetic} has proposed to rely on intuitively meaningful edits that are represented by conventional post-processing operations, i.e., image filters, to make the automatic process more understandable to users.  
Moreover, such methods have much fewer parameters to optimize, and can be applied resolution-independently.

\noindent\textbf{Problem Formulation.}
A neural network can be denoted as a function $F(\boldsymbol{x})=\boldsymbol{y}$ that outputs $\boldsymbol{y}\in\mathbb{R}^m$ for an image $\boldsymbol{x}\in\mathbb{R}^n$.
Here we focus on the widely-used DNN classifier with a softmax function, which expresses the output $\boldsymbol{y}$ as a probability distribution, i.e., $0\leq\boldsymbol{y}_i\leq1$ and $ \boldsymbol{y}_1+\cdot\cdot\cdot+\boldsymbol{y}_m=1$.
The final predicted label $l$ for $\boldsymbol{x}$ is accordingly obtained by $l={\mathrm{arg max}}_i~\boldsymbol{y}_i$.
An adversary aims to induce a misclassification of a DNN classifier $F(\boldsymbol{x})$ through modifying the original image $\boldsymbol{x}$ into $\boldsymbol{x}'$ such that $F(\boldsymbol{x}')\neq{y}$.

\noindent\textbf{Restricted Adversary with Imperceptible Perturbations.}
As mentioned in Section~\ref{sec:intro}, in order to make the modification unrecognizable, most existing work forces the adversarial image $\boldsymbol{x}'$ to be visually close to its original image $\boldsymbol{x}$ with respect to specific distance measurements.
The conventional solution is $L_{p}$ distance (typically $L_{\infty}$~\cite{carlini2017towards,goodfellow2014explaining,kurakin2016adversarial,madry2017towards} and $L_{2}$~\cite{carlini2017towards,moosavi2016deepfool,rony2019decoupling,szegedy2013intriguing}, but also $L_{1}$~\cite{chen2018ead} and $L_{0}$~\cite{papernot2016limitations,su2019one}).
The earliest work in this direction~\cite{szegedy2013intriguing}
proposed to jointly optimize   misclassification with cross-entropy loss and the $L_{2}$ distance by solving a box-constrained optimization with the L-BFGS method~\cite{liu1989limited}.
The C\&W method~\cite{carlini2017towards} followed a similar idea, but replaced the cross-entropy loss with another specially designed loss function, namely, the differences between the pre-softmax logits.
Moreover, a new variable was introduced to eliminate the box constraint. 
The method can be expressed as:

\begin{equation}
\label{cw}
\begin{gathered}
\underset{\boldsymbol{w}}{\mathrm{minimize}}
~~{{\|\boldsymbol{x}'-\boldsymbol{x}\|}_2^2}+\lambda f(\boldsymbol{x}'),\\
\textrm{where}~~ f(\boldsymbol{x}')=\mathrm{max}(Z(\boldsymbol{x}')_l-{\mathrm{max}}\{Z(\boldsymbol{x}')_i:i\neq l\},-\kappa),\\
\textrm{and}~~\boldsymbol{x}'=\frac{1}{2}(\tanh(
\mathrm{arctanh}(\boldsymbol{x})+\boldsymbol{w})+1),
\end{gathered}
\end{equation}
where $f(\cdot)$ is the new loss function, $\boldsymbol{w}$ is the new variable, and $Z(x')_i$ is the logit with respect to the $i$-th class given the intermediate modified image $\boldsymbol{x}'$.
The parameter $\kappa$ is applied to control the confidence level of the misclassification.

This joint optimization is straightforward but suffers from high computational cost due
to the need for line search to optimize $\lambda$.
For this reason, other methods~\cite{goodfellow2014explaining,dong2018boosting,kurakin2016adversarial,madry2017towards,rony2019decoupling} instead rely on Projected Gradient Descent (PGD) to restrict the perturbations with a small $L_p$ norm bound, $\epsilon$.
Specifically, the fast gradient sign method (FGSM)~\cite{goodfellow2014explaining} was designed to succeed within only one step 
and was extended by~\cite{dong2018boosting,kurakin2016adversarial,madry2017towards,rony2019decoupling} to exploit finer gradient information with multiple iterations.
The iterative approach can be formulated as:
\begin{equation}
\label{IFGSM}
% \begin{gathered}
{\boldsymbol{x}_0'}=\boldsymbol{x},~~{\boldsymbol{x}_{k}'}={\boldsymbol{x}_{k-1}'}+\alpha\cdot\ \mathrm{sign}({\nabla_{{\boldsymbol{x}}}J(\boldsymbol{x}_{k-1}',l)}),
% \end{gathered}
\end{equation}
%%adding the bound restriction
where $\alpha$ denotes the step size in each iteration.
The generated adversarial perturbations will be clipped to satisfy the $L_{\infty}$ bound.
A generalization of this formulation to the $L_2$ norm can be achieved by replacing the $\mathrm{sign} (\cdot)$ with a normalization operation~\cite{moosavi2016deepfool,rony2019decoupling}.
$L_0$ and $L_1$-bounded adversarial images were also studied~\cite{chen2018ead,papernot2016limitations,su2019one}, but not widely adopted since the resulting sparse perturbations are not stable in practice.

Recently, there have also been several attempts to address the limitations of naive $L_p$ by using more perception-aligned solutions for measuring similarity.
A straightforward way is by incorporating existing metrics, such as Structural SIMilarity (SSIM)~\cite{rozsa2016adversarial}, Wasserstein distance~\cite{wong2019wasserstein}, and the perceptual color metric CIEDE2000~\cite{zhao2020towards}.
Other methods~\cite{Croce_2019_ICCV,luo2018towards,zhang2020smooth} adapted the $L_p$ measurements to the textural properties of the image, i.e., hiding perturbations in image regions with high visual variation.
Local pixel displacement was also explored~\cite{xiao2018spatially,alaifari2018adef}. 
In general, these solutions yield a better trade-off between adversarial strength and
imperceptibility than conventional $L_p$ methods.

\noindent\textbf{Unrestricted Adversaries towards Realistic Images.}
Due to the assumption of imperceptible perturbations, now considered unrealistic, as mentioned in Section~\ref{sec:intro}, recent work has started to pursue non-suspicious adversarial images with large perturbations, which make more sense in practical use scenarios.
Common approaches to creating such unrestricted adversarial images can be divided into three categories: geometric transformation, semantic manipulation, and color modification.
The geometric transformation method penalizes image differences with respect to small
rotations and translations of the image~\cite{engstrom2019exploring}.
Semantic manipulation has been so far mainly studied in the domain of face recognition, where the perturbation is optimized with respect to specific semantic attribute(s),
such as colors of skin and extent of makeup~\cite{joshi2019semantic,qiu2019semanticadv,sharif2019adversarial}.

Existing colorization-based work has explored
uniform color transformation~\cite{hosseini2018semantic,Laidlaw2019functional,shamsabadi2020colorfool} and
automatic colorization~\cite{bhattad2020Unrestricted}.
Specifically, the early method~\cite{hosseini2018semantic} randomly adjusts the hue values of each image pixel to
search for possible adversarial images.
The ColorFool method~\cite{shamsabadi2020colorfool} improves on~\cite{hosseini2018semantic} by imposing semantic-aware norm constraints for better image quality, but still relies on costly random search.
The ReColorAdv method~\cite{Laidlaw2019functional} optimizes color transformation over a discretely parameterized color space with post-interpolation and regularization on local uniformity, and impose $L_{\infty}$ bounds on
the perturbations.
The cAdv method~\cite{bhattad2020Unrestricted} takes a different route, optimizing a pre-trained automatic colorization model to re-colorize the gray-scale version of the original image.
It increases the computational overhead
due to the huge number of parameters in the deep colorization model, and also has been shown to cause abnormal color stains (see examples in ~\cite{bhattad2020Unrestricted} and our Figure~\ref{fig:compare_un}).

Our ACE falls into the colorization category
but is markedly different from existing approaches.
Specifically, ACE creates adversarial images by optimizing with gradient information, and, in this way, is fundamentally different from the random search-based approaches in~\cite{hosseini2018semantic,shamsabadi2020colorfool}.
In Section~\ref{sec:exp}, we also show that our gradient-based ACE outperforms its alternative with random search.
Our color filter is simpler and more transparent than the deep colorization model in \cite{bhattad2020Unrestricted}.
Compared with~\cite{Laidlaw2019functional}, our ACE enjoys a more elegant and continuous formulation.
Experimental results in Section~\ref{sec:exp} demonstrate that our ACE outperforms these approaches in both adversarial strength and image quality.

\section{Adversarial Color Enhancement (ACE)}
This section describes our proposed Adversarial Color Enhancement (ACE),
which generates visually realistic adversarial filtered images based on a commonly-used color filter.
Specifically, we adopt the differentiable approximation in ~\cite{hu2018exposure} to parameterize the color filter by a monotonic piecewise-linear mapping function with totally $K$ pieces:
\begin{equation}
\label{filter}
\begin{gathered}
F_{\boldsymbol{\theta}}(x_k)=\sum\limits_{i=1}^{k-1}{\theta_i}+(K \cdot x_k-(k-1)) \cdot \theta_k ,\\
\textrm{s.t.}~0\leq\theta_i\leq1~\mathrm{and}~\sum_i \theta_i=1,
\end{gathered}
\end{equation}
where $x_k$ denotes any image pixel whose value falls into the $k$-th piece of the mapping function, and $F_{\boldsymbol{\theta}}(x_k)$ is its corresponding output after filtering.
An example of this function with four pieces ($K=4$) is illustrated in Figure~\ref{fig:examples} (right).

Note that we are not optimizing in the
pixel space but in the latent space of filter parameters, and the three RGB channels are operated on in parallel.
The parameters $\boldsymbol{\theta}$ ($K$ in total) can be optimized via gradient descent to achieve a specific objective.
Obviously, an image will remain unchanged ($F_{\boldsymbol{\theta}}(x)=x$) when all the parameters are equal to $1/K$.
As a result, we propose to control over the adjustment by imposing constraints on the distance between each parameter and its initial value $1/K$.
The misclassification objective and the proposed constraints on the parameters will be jointly optimized with a balance factor $\lambda$, expressed as: 
\begin{equation}
\label{opt_ACE}
\underset{\boldsymbol{\theta}}{\mathrm{minimize}}~f(F_{\boldsymbol{\theta}}(\boldsymbol{x})) +\lambda \cdot\sum_i({\theta_i}-1/K)^2,
\end{equation}
where $f(\cdot)$ is the C\&W loss on logit differences in Equation~\ref{cw}.
% \vspace{-0.3cm}

\begin{table}[b]
\renewcommand{\arraystretch}{1}
    \begin{minipage}{.6\textwidth}
      \centering
      \resizebox{\textwidth}{!}{
        \begin{tabular}{l|ccccc}
\toprule[1pt]
&$\rightarrow$Alex& $\rightarrow$R50&$\rightarrow$V19&$\rightarrow$D121&$\rightarrow$Inc3\\
\midrule[1pt]

Alex&99.9&6.26&7.10&6.85&2.08\\
R50&48.50&98.3&15.83&13.21&5.96\\
V19&39.52&11.29&98.5&10.65&10.30\\
D121&46.50&18.51&15.16&98.4&5.61\\
Inc3&41.12&16.42&14.87&12.35&93.2\\
\bottomrule[1pt]
\end{tabular}}
    \end{minipage}%
    \begin{minipage}{.4\textwidth}
      \centering
        \begin{tabular}{l|ccc}
\toprule[1pt]
&$\rightarrow$Alex&$\rightarrow$R18&$\rightarrow$R50\\
\midrule[1pt]
Alex&100.0&16.40&12.30\\
R18&48.52&99.2&22.43\\
R50&48.98&30.47&99.0\\
\bottomrule[1pt]
\end{tabular}
    \end{minipage} 
         \caption{White-box success rates (diagonal) and black-box transferability of our ACE in ImageNet classification (\textbf{left}) and private scene recognition (\textbf{right}).
         Considered models: AlexNet (Alex), ResNet18 (R18), ResNet50 (R50), VGG19 (V19), DenseNet121 (D121), and Inception-V3 (Inc3). The success rates are measured with respect to the models in the columns when applying ACE on the models in the rows.}
         \label{tab:cross}
\end{table}

\section{Experiments}
\label{sec:exp}
We evaluate our ACE in two different tasks: object classification and scene recognition, and consider
the following two datasets.
\noindent\textbf{ImageNet-Compatible dataset} consists of 6000 images associated with ImageNet class labels, and has been used in the NIPS 2017 Competition on Adversarial Attacks and Defenses~\cite{kurakin2018adversarial}.
Here we use its development set containing 1000 images.
\noindent\textbf{Private Scene dataset} was introduced by the MediaEval Pixel Privacy task~\cite{ppoverview2018}, which aims to develop image modification techniques that help to protect users against automatic inference of privacy-sensitive scene information.
It contains 600 images with 60 privacy-sensitive scene categories, selected from the Places365 dataset~\cite{zhou2017places}.
For the ImageNet task, we consider five distinct classifiers that are pre-trained on ImageNet: AlexNet~\cite{krizhevsky2012imagenet}, ResNet50~\cite{he2016deep}, VGG19~\cite{simonyan2014very}, DenseNet121~\cite{huang2017densely}, and Inception-V3~\cite{szegedy2016rethinking}.
For the scene task, we consider AlexNet, ResNet18, and ResNet50 pre-trained on the Places365 dataset.

Our experiments are carried out on a single NVIDIA Tesla P100 GPU with 12 GB of memory. ACE is optimized using Adam~\cite{kingma2014adam} with a learning rate of 0.01, under a maximum budget of 500 iterations.
We execute the optimization in 40 batches of 25 image samples, and the run-time on ImageNet task is about 2 seconds per image.
Early stopping is triggered when the optimization is no longer making progress as implemented in~\cite{carlini2017towards,rony2019decoupling}.
If not mentioned specifically, ACE is implemented with the optimal settings, $K=64$ and $\lambda=5$.
Table~\ref{tab:cross} shows that our ACE can achieve high white-box success rates and have good cross-model transferability.
It can also be observed that models with more sophisticated architecture are generally harder to fool in the white-box case, and transferring from a sophisticated architecture to a simple one is easier than the other way around.
Note that the transferability is calculated on images for which the prediction of both the models involved is the same.

\begin{table}[h]
\newcommand{\tabincell}[2]{\begin{tabular}{@{}#1@{}}#2\end{tabular}}
\renewcommand{\arraystretch}{1}
\begin{center}
\resizebox{\textwidth}{!}{
\begin{tabular}{l|ccc|c|cccc}
\toprule[1pt]
&\multicolumn{3}{c|}{$L_p$ Norm}&\multicolumn{5}{c}{Success Rate}\\
\hline
&$L_0$ (\%)&$L_2$&$L_{\infty}$&Inc3&$\rightarrow$Alex& $\rightarrow$R50&$\rightarrow$V19&$\rightarrow$D121\\
\midrule[1pt]
FGSM~\cite{goodfellow2014explaining}&49.34&4.05&2.00&78.10&7.84&5.40&5.74&5.50\\
BIM~\cite{kurakin2016adversarial} &39.23&3.09&2.00&99.1&8.16&4.95&6.44&4.71\\
C\&W~\cite{carlini2017towards}&29.06&3.00&15.66&99.6&8.16&4.72&6.79&4.38\\
\midrule[1pt]
ReColorAdv~\cite{Laidlaw2019functional}&70.81&18.87&64.00&79.3&9.76&4.50&3.40&2.58\\
ReColorAdv$^+$~\cite{Laidlaw2019functional}&82.50&47.53&97.21&89.2&31.20&15.64&13.58&10.77\\
cAdv~\cite{bhattad2020Unrestricted}&41.42&20.54&116.15&91.8&30.08&11.25&11.01&\textbf{13.47}\\
Our ACE&42.99&40.61&45.98&93.2&\textbf{41.12}&\textbf{16.42}&\textbf{14.87}&12.35\\
\bottomrule[1pt]
\end{tabular}
    }
        
\end{center}
% \vspace{-0.3cm}
\caption{White-box success rates and black-box transferability of our ACE compared with other baselines.
$L_0$ is the proportion of the perturbed pixels and $L_{\infty}$ is shown in [0,255]. Here the Inception-V3 is used as the target white-box model.}
\label{tab:trans}
% \vspace{-0.4cm}
\end{table}

\subsection{Comparisons on Adversarial Strength and Image Quality}
We further compare ACE with the following gradient-based baseline methods in terms of adversarial strength and image quality, in the ImageNet task:

\noindent\textbf{FGSM}~\cite{goodfellow2014explaining} with a $L_{\infty}$ norm bound $\epsilon=2/255$ for ensuring imperceptibility.

\noindent\textbf{BIM}~\cite{kurakin2016adversarial} with a $L_{\infty}$ norm bound $\epsilon=2/255$, and 10 iterations of gradient descent.

\noindent\textbf{C\&W}~\cite{carlini2017towards} optimized on $L_2$ with fewer iterations and higher confidence level (iters=$3\times100$ and $\kappa=40$) than usual to yield larger perturbations for stronger adversarial effects.

\noindent\textbf{ReColorAdv}~\cite{Laidlaw2019functional} (Unrestricted) with $\epsilon=16/255$ and lr=0.001 as in~\cite{Laidlaw2019functional},
and another version allowing larger perturbations ($\epsilon=51/255$ and lr=0.005), denoted as ReColorAdv$^+$.

\noindent\textbf{cAdv}~\cite{bhattad2020Unrestricted} (Unrestricted) with the settings leading to optimal color realism ($k=8$).
Note that cAdv can only produce adversarial images sized $224\times224$ due to the fixed output resolution of its pre-trained deep colorization model.

\begin{figure}[h]
\begin{center}
% \fbox{\rule{0pt}{3.2in} \rule{\textwidth}{0pt}}
  \includegraphics[width=0.9\textwidth]{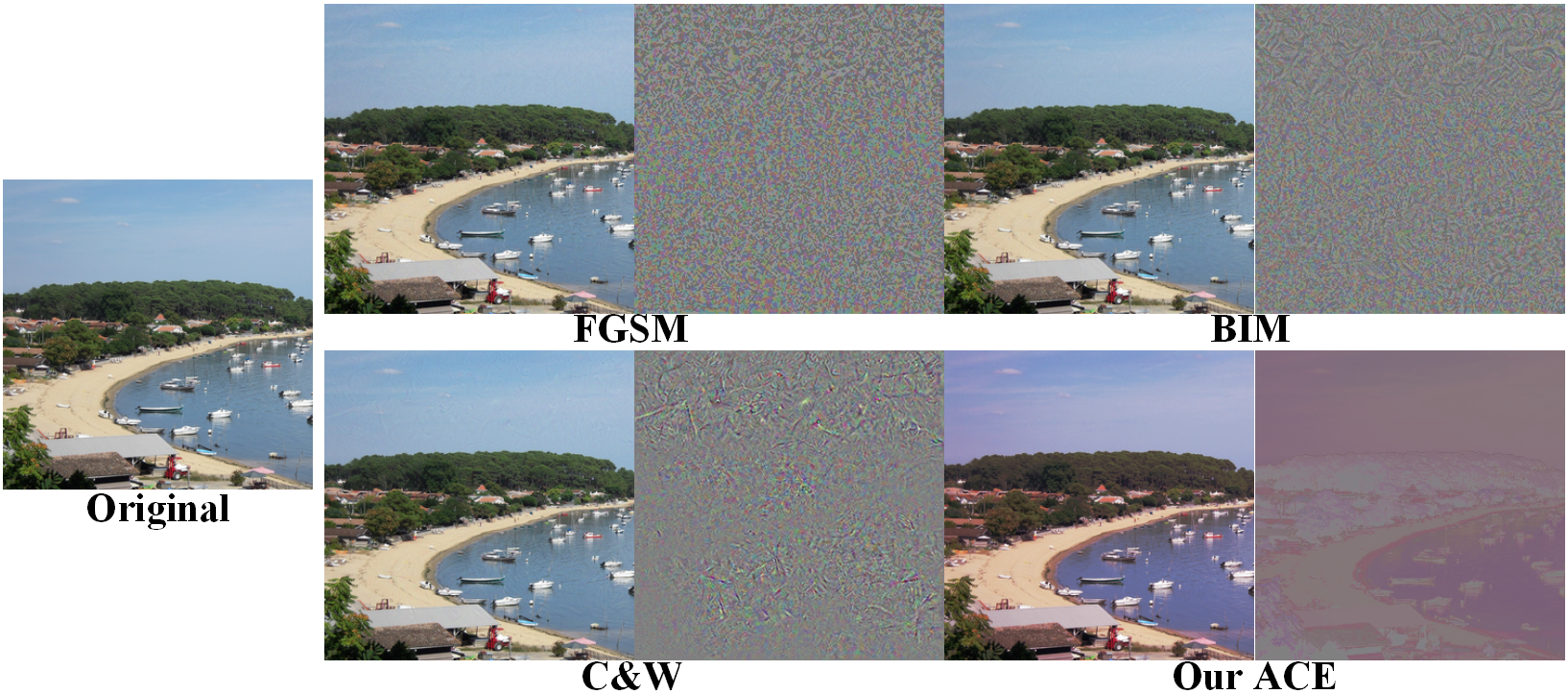}
\end{center}
\vspace{-0.2cm}
   \caption{Adversarial images and perturbations by our ACE and $L_p$ methods FGSM~\cite{goodfellow2014explaining}, BIM~\cite{kurakin2016adversarial} and C\&W~\cite{carlini2017towards}. ACE yields more natural appearances without abnormal patterns.}
\label{fig:compare_lp}
\vspace{-0.2cm}
\end{figure}

\begin{figure}[h]
\begin{center}
% \fbox{\rule{0pt}{3.2in} \rule{\textwidth}{0pt}}
  \includegraphics[width=0.8\textwidth]{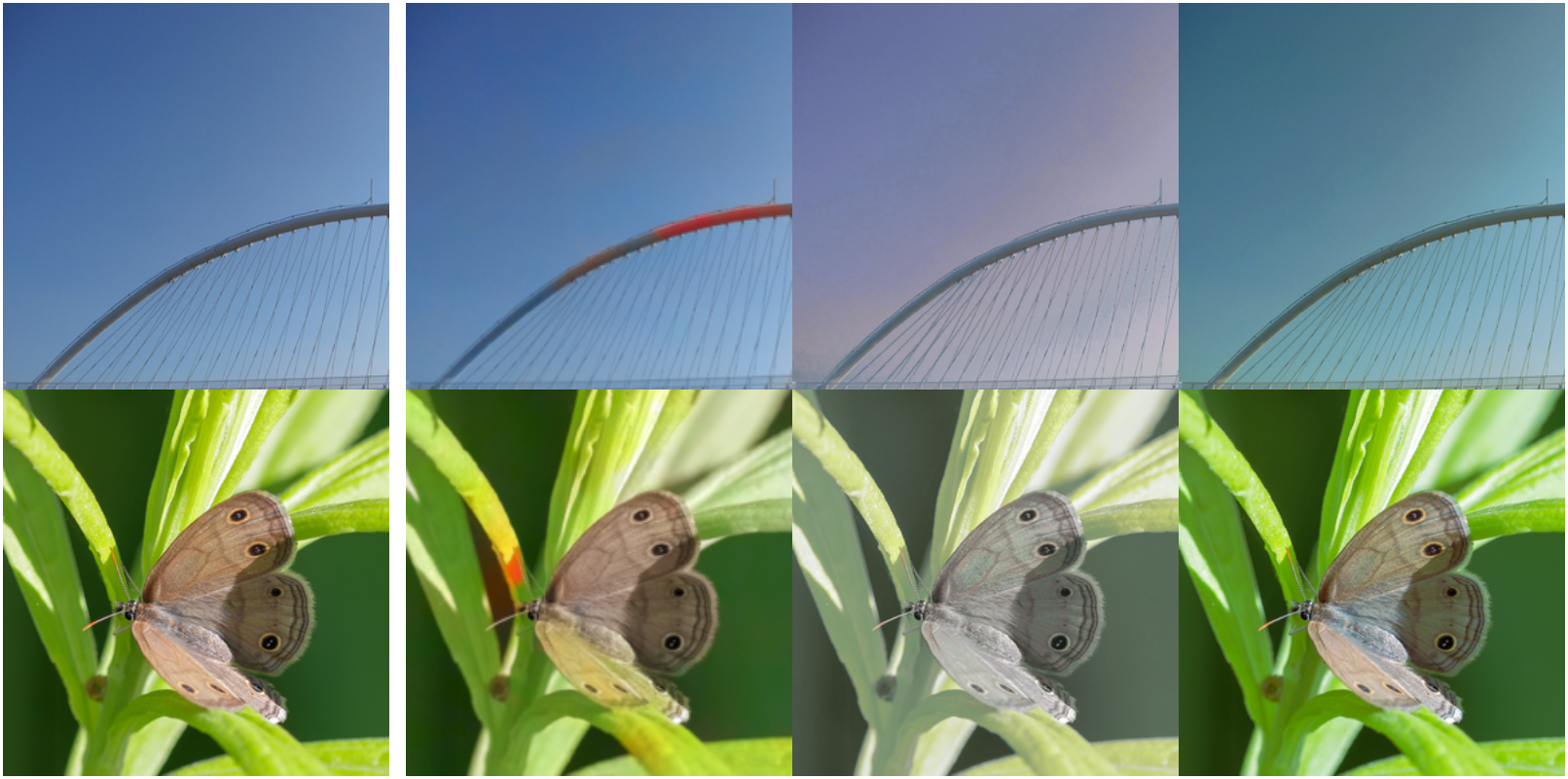}
\end{center}
\vspace{-0.2cm}
   \caption{Examples from three different unrestricted methods. From left to right: original image, adversarial images generated by cAdv~\cite{bhattad2020Unrestricted}, ReColorAdv$^+$~\cite{Laidlaw2019functional}, and our ACE. ACE yields more acceptable images with smooth appearances and fewer artifacts.}
\label{fig:compare_un}
\vspace{-0.2cm}
\end{figure}

We adopt Inception-V3 as the white-box model because it is the official model used in the NIPS 2017 Competition.
As shown in Table~\ref{tab:trans}, ACE can consistently achieve better transferability than conventional $L_p$ methods, while not 
introducing visually suspicious noisy patterns (see Figure~\ref{fig:compare_lp}).
Iterative $L_p$ methods (BIM and C\&W) could achieve the strongest white-box adversarial effects by fully leveraging the gradient information, but these effects are less generalizable to other unseen models, i.e., worse transferability.
Among the unrestricted methods, our ACE achieves the highest white-box success rates and overall best transferability, while 
yielding smooth adjustment without abnormal colorization artifacts (see Figure~\ref{fig:compare_un}).
Such smoothness is also reflected in $L_{\infty}$ norms that are lower than other unrestricted methods, meaning that ACE tends to avoid excessive local color changes.

\subsection{Ablation Study}
\begin{figure}[t]
\centering
  \includegraphics[width=\textwidth]{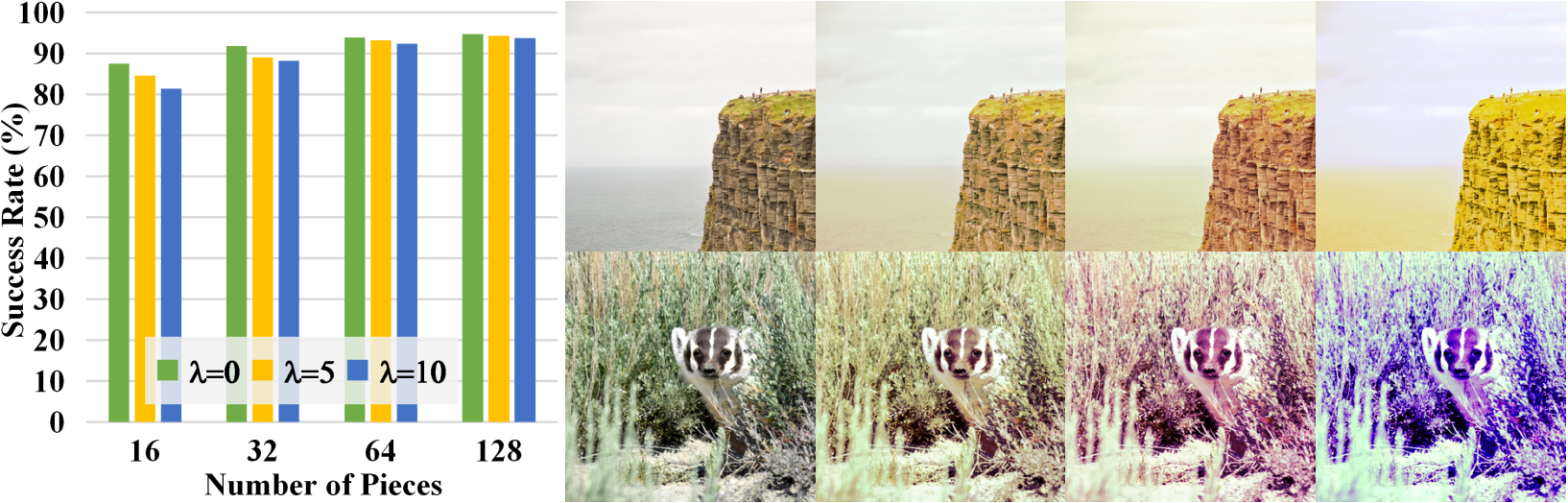}
   \caption{\textbf{Left:} White-box success rates of ACE when varying the number of pieces $K$ with four different $\lambda$ values. \textbf{Right:} Adversarial images with varied $\lambda$. For each example, from left to right: original image, adversarial images with $\lambda=$10, 5 and 0. Too large a $\lambda$ makes ACE not reach its full potential, while setting $\lambda=0$ may cause unrealistic colorization.}
\label{fig:k_lambda}
\vspace{-0.2cm}
\end{figure}

\noindent\textbf{Hyperparameters.} Figure~\ref{fig:k_lambda} (left) shows the success rates of ACE with a different number of pieces $K$ under different factor $\lambda$ values used for balancing the two loss terms in the joint optimization.
We can observe that increasing $K$ slightly improves the performance by expanding the action space of the adversary.
Moreover, increasing $K$ allows more fine-grained color adjustment in the images that have rich colors.
It should also be noted that using more pieces means more computational cost during the optimization.

On the other hand, relaxing the constraints by decreasing $\lambda$ gives the adversary larger action space, leading to higher success rates.
However, completely removing the constraints ($\lambda=0$)  will lead to unrealistic image appearances, as can be observed in Figure~\ref{fig:k_lambda} (right).
Specifically, in this paper, we use $K=64$ and $\lambda=5$ as optimal settings for a good trade-off.

\noindent\textbf{Gradient Descent vs. Random Search.}
We compare our gradient-based ACE with a random search-based implementation using the same color filter. 
In this case, the parameters will be updated with gradient information in our ACE, while being uniformly sampled from the valid range [0,1] for random search.

\begin{figure}[b]
\centering
\vspace{-0.2cm}

  \includegraphics[width=0.9\columnwidth]{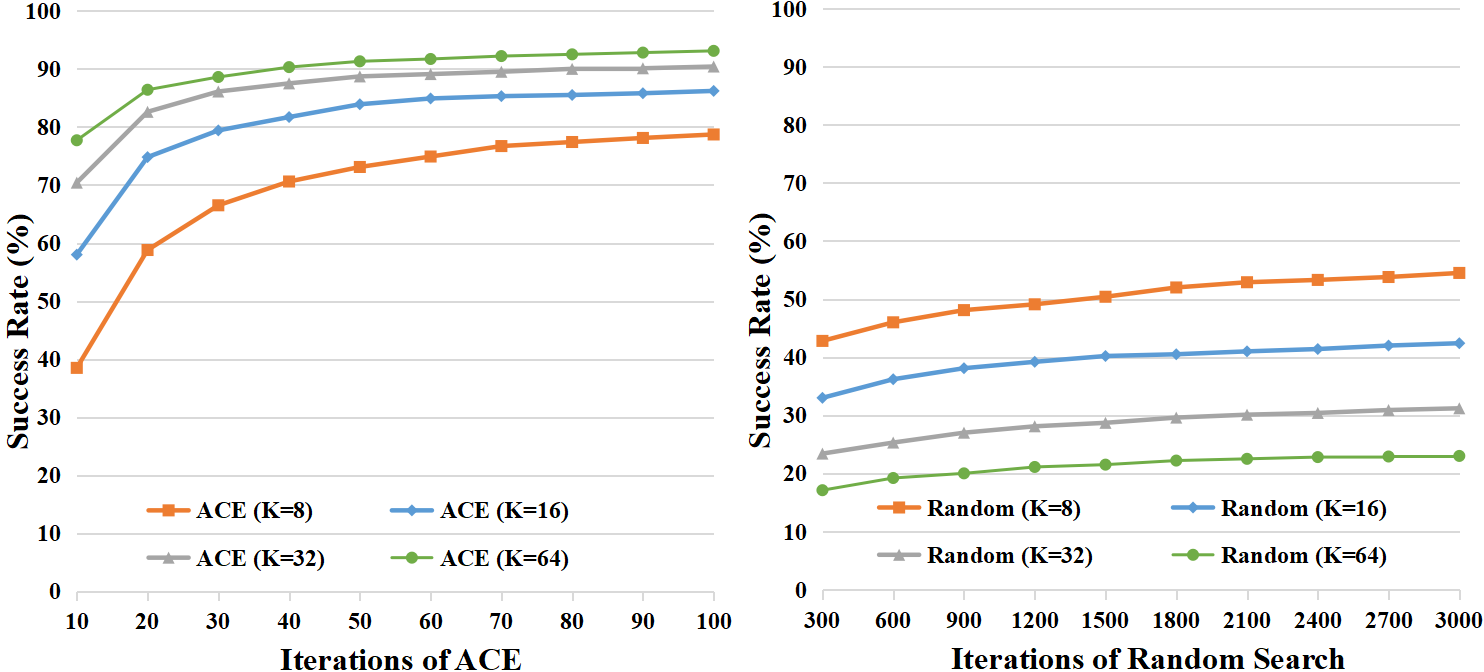}
   \caption{White-box success rates for our ACE (\textbf{left}) and random search (\textbf{right}) as a function of maximum allowed iterations. Note that the scales on the two x-axes are not the same.}
\label{fig:ace_vs_rand}
\vspace{-0.2cm}
\end{figure}

Figure~\ref{fig:ace_vs_rand} shows their white-box success rates as a function of iterations.
For random search, we repeat several times and got almost the same results.
We can observe that our ACE consistently outperforms random search, even with far fewer iterations.
Moreover, ACE gradually improves as the number of parameters $K$ increases, indicating that it can benefit from the expanded action space for more fine-grained color adjustment.
In contrast, random search becomes increasingly worse since the potentially successful adversarial samples can no longer be feasibly found in the exponentially expanded parameter space.
For comparison we mention that, the grid search-based ColorFool~\cite{shamsabadi2020colorfool} has a success rate of 64.6\% (65.4\%) with 1000 (1500) iterations when testing on our dataset with the Inception-V3 as the white-box model. 

\begin{figure}[t]
\begin{center}
  \includegraphics[width=0.7\columnwidth]{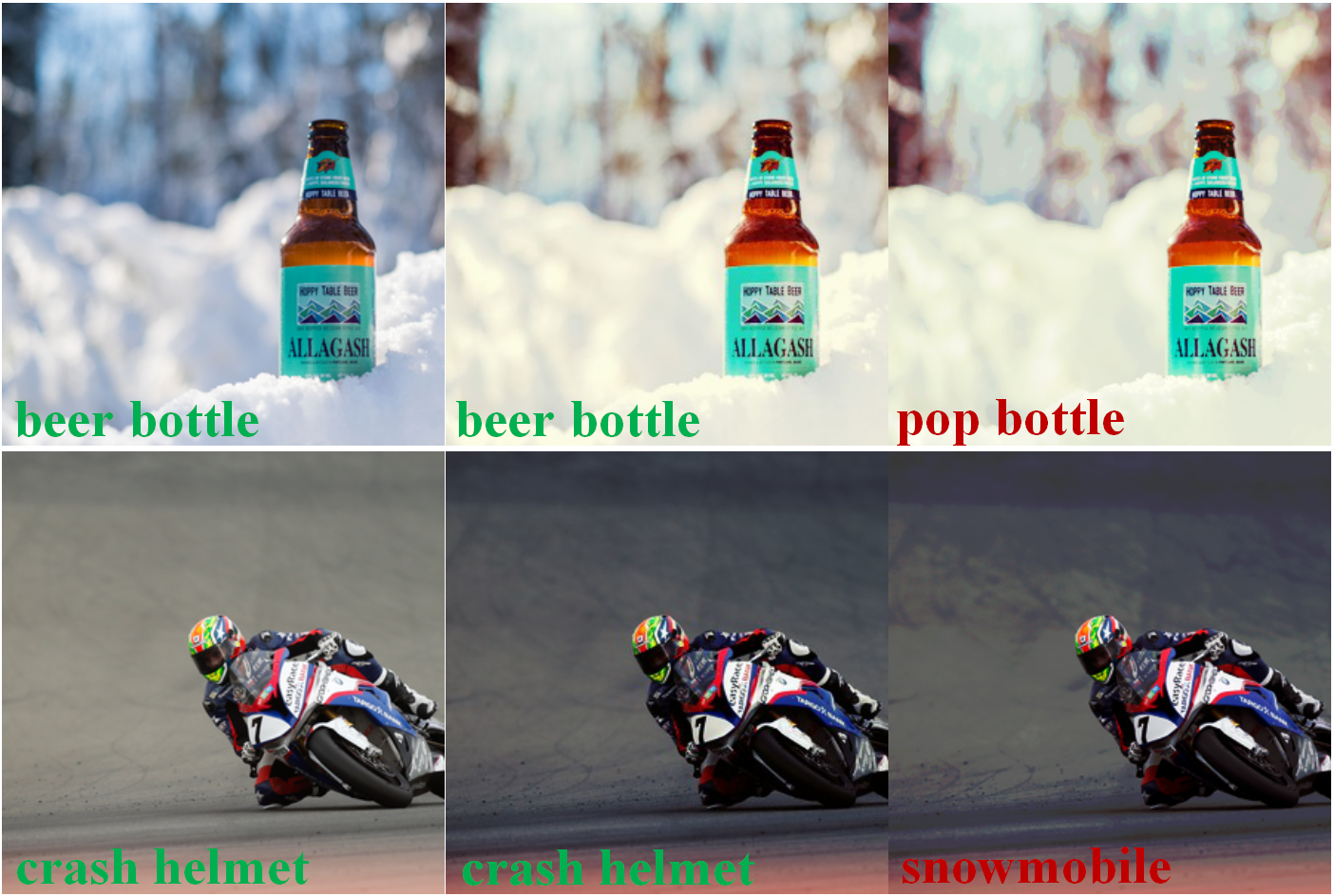}
\end{center}
% \vspace{-0.2cm}
   \caption{Adversarial filtered images by ACE with guidance of specific image filters. Two Instagram filters are considered: Hefe (\textbf{top}) and Gotham (\textbf{bottom}). For each, from left to right: original image, Instagram filtered image and adversarial filtered image by ACE.}
\label{fig:ins}
% \vspace{-0.2cm}
\end{figure}

\begin{figure}[b]
\begin{center}
% \fbox{\rule{0pt}{3.2in} \rule{\textwidth}{0pt}}
  \includegraphics[width=\columnwidth]{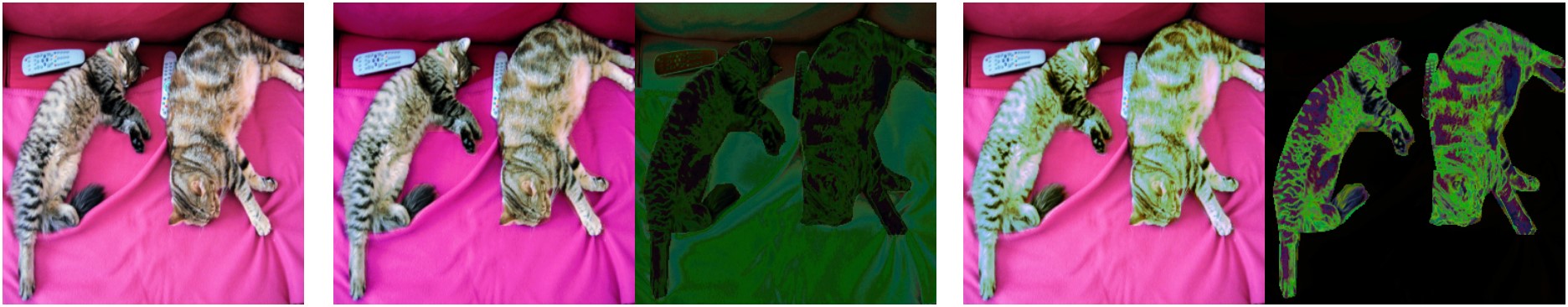}
\end{center}
% \vspace{-0.1cm}
   \caption{Adapting ACE on different semantics. Original image (\textbf{left}), successful adversary with stronger constraints for cat but weaker for blanket (\textbf{middle}), or vice versa (\textbf{right}). The middle case yields more realistic results since blanket naturally occurs with various colors.}
\label{fig:adapt}
% \vspace{-0.2cm}
\end{figure}

\section{ACE Extensions}
\noindent\textbf{Adaption on Image Styles.}
Previous work~\cite{choi2017geo} has pointed out that popular image enhancement practices can potentially degrade automatic inference.
Despite our ACE in Equation~\ref{opt_ACE}
achieving visually acceptable results, it is not directly optimized towards enhancing image quality.
Therefore, we explore the possibility to guide ACE towards achieving quality enhancement in addition to the adversarial effects. 
Specifically, we propose to optimize the adversarial image towards specific attractive styles that were obtained by using Instagram filters.
Accordingly, the optimization objective is adjusted to: 
\begin{equation}
\label{opt_ins}
\underset{\boldsymbol{\theta}}{\mathrm{minimize}}~f(F_{\boldsymbol{\theta}}(\boldsymbol{x})) +\lambda \cdot{\|F_{\boldsymbol{\theta}}(\boldsymbol{x})-\boldsymbol{x}_t\|}_2^2,
\end{equation}
where $\boldsymbol{x}_t$ denotes the target Instagram filtered image, and the final adversarial filtered image $F_{\boldsymbol{\theta}}$ is therefore guided to have similar appearances with $\boldsymbol{x}_t$ by minimizing their distance.
As shown in Figure~\ref{fig:ins}, this adaptation of ACE can successfully enhance the image by mimicking the effects of Instagram filters.  

\noindent\textbf{Adaptation on Semantics.}
ACE treats all the image pixels that have the same values in the same way.
Inspired by previous work~\cite{bhattad2020Unrestricted,shamsabadi2020colorfool}, we show that semantically adapting ACE could better maintain image quality by hiding large perturbations in the semantic regions that remain realistic with various colors.
Specifically, Equation~\ref{opt_ACE} is adapted to:
% \begin{equation}
% \label{opt-adv}
% % \begin{gathered}
% \underset{\boldsymbol{\theta}}{\mathrm{minimize}}~\cdot J(F_{\boldsymbol{\theta}_n}(\boldsymbol{x} \cdot M_n))+ \lambda \sum_{n,i} w_n \cdot({\theta_{n_i}}-1/K)^2,~\textrm{s.t.}~\sum_n w_n=1,
% % \end{gathered}
% \end{equation}
\begin{equation}
\label{opt-adv}
% \begin{gathered}
\underset{\boldsymbol{\theta}}{\mathrm{min}}~\sum_{n} [J(F_{\boldsymbol{\theta}_n}(\boldsymbol{x} \cdot M_n))+ \lambda \sum_{i} w_n \cdot({\theta_{n_i}}-1/K)^2],~\textrm{s.t.}~\sum_n w_n=1,
% \end{gathered}
\end{equation}
where $w_n$ is the weight for the $n$-th filter, $F_{\boldsymbol{\theta}}^{n}(\cdot)$, which is optimized independently for a specific semantic region given its mask $M_n$ obtained by a semantic segmentation method~\cite{kirillov2019panoptic}.
As shown in Figure~\ref{fig:adapt}, this adaption avoids raising the sense of unrealistic colorization, leading to improved image quality.

\section{Conclusion}
We have proposed Adversarial Color Enhancement (ACE), an approach to generating unrestricted adversarial images by optimizing a color filter via gradient descent.
ACE has been shown to produce realistic filtered images with good transferability, which results in strong real-world black-box adversaries.
We also present two potential ways to improve ACE in terms of image quality by guiding it with specific attractive image styles or adapting it to regional semantics.

In the current ACE, the single hyperparameter of the filter ($K$ in Equation~\ref{filter}) is per-fixed for all images without considering their individual properties.
Since natural images would differ in the range and complexity of their contained colors, adaptive strategies would be worth exploring in order to yield more suitable modifications.
For example, images with most pixels concentrated in a certain color range should have larger action space in that range than those with more uniform color distribution.
It would also be interesting to carry out user study on the visual quality of the images generated by ACE.
On the other side, developing defenses against the proposed adversarial color filtering is necessary to make current neural networks more robust, based on either adversarial training or algorithms for identifying adversarial modifications by ACE.
\section*{Acknowledgement}
This work was carried out on the Dutch national e-infrastructure with the support of SURF Cooperative.
% \clearpage
% \newpage
\bibliography{ref}

\end{document}